\documentclass{svproc}
\usepackage{hyperref}
\usepackage{xcolor}
\usepackage{amsmath}
\usepackage{graphicx}
\usepackage{booktabs}
\usepackage{url}

\usepackage{diagbox}
\usepackage{subcaption}
\usepackage{multirow}
\usepackage{placeins}
\hypersetup{colorlinks, citecolor=red,linkcolor=red,urlcolor=blue}

\begin{document}
\mainmatter              
\title{Towards the Localisation of Lesions in Diabetic Retinopathy}
\titlerunning{Towards Lesion Localisation}  
%
\author{Samuel Ofosu Mensah\inst{1, 2},  Bubacarr Bah\inst{1, 2} \and
Willie Brink\inst{2}}
%

\institute{African Institute for Mathematical Sciences, Cape Town, South Africa \\
	\texttt{samuelmensah@aims.ac.za}
\and
Department of Mathematical Sciences, Stellenbosch University, South Africa}

\maketitle              

\begin{abstract}\fontsize{9}{11}\selectfont{
Convolutional Neural Networks (CNNs) have successfully been used to classify diabetic retinopathy (DR) fundus images in recent times. However, deeper representations in CNNs may capture higher-level semantics at the expense of spatial resolution. To make predictions usable for ophthalmologists, we use a post-attention technique called Gradient-weighted Class Activation Mapping (Grad-CAM) on the penultimate layer of deep learning models to produce coarse localisation maps on DR fundus images. This is to help identify discriminative regions in the images, consequently providing evidence for ophthalmologists to make a diagnosis and potentially save lives by early diagnosis. Specifically, this study uses pre-trained weights from four state-of-the-art deep learning models to produce and compare localisation maps of DR fundus images. The models used include VGG16, ResNet50, InceptionV3, and InceptionResNetV2. We find that InceptionV3 achieves the best performance with a test classification accuracy of 96.07\%, and localise lesions better and faster than the other models. 
}
\end{abstract}

\section{Introduction}
There has been tremendous success in the field of deep learning, especially in the computer vision domain. This success may be attributed to the invention of Convolutional Neural Networks (CNNs). It has since been extended to specialised fields such as medicine \cite{raghu}. With the help of transfer learning \cite{tan}, pre-trained weights of state-of-the-art models can be used for the analysis of medical images. In this study, for example, we use deep learning backed with transferred weights of pre-trained models to classify Diabetic Retinopathy (DR) fundus images \cite{gulshan,raghu}. Even though these models are able to achieve good performance, it can be difficult to understand the reasoning behind their discrimination processes. This is due in part to the fact that deep learning models have a nonlinear multilayer structure \cite{gondal}. 

Recently, DR classification has received a lot of attention as it has been useful in the ophthalmology domain \cite{gulshan,raghu}. However, ophthalmologists may not be able to evaluate the true performance of the model \cite{gondal,beede}. We attempt to mitigate this issue by providing additional evidence of model performance \cite{gondal} with the help of a post-attention technique called Gradient-weighted Class Activation Mapping (Grad-CAM) \cite{selvaraju}. This technique can be seen as a Computer-Aided Diagnosis (CAD) tool integrated to increase the speed of diabetic retinopathy diagnosis.

The main contribution of this paper is to use Grad-CAM to generate coarse localisation maps from higher-level semantics of a CNN model, consequently aiding and speeding up the diagnosis process of diabetic retinopathy.

The rest of this paper is organized as follows. In Section \ref{s2}, we introduce detail of the data and techniques used for the study. Section \ref{s3} provides performances of the evaluation done for the models used in the study. Finally, we present conclusions and future work in Section \ref{s4}.


\section{Data and Methodology}\label{s2}
\begin{figure}[h]
	\centering
	\framebox{\includegraphics[scale=.5]{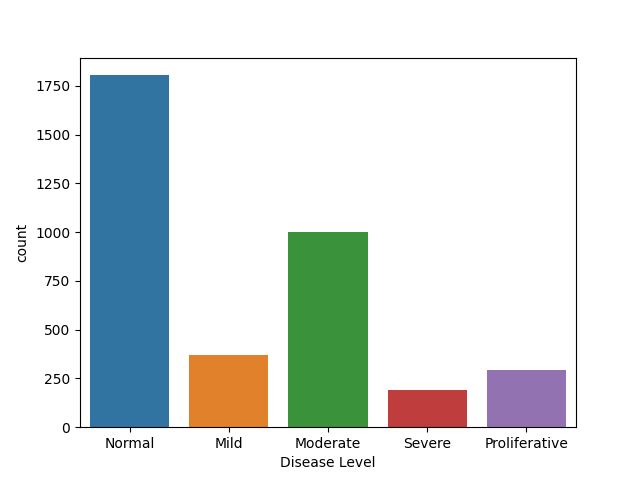}}
	\caption{The distribution of classes for APTOS dataset.}
	\label{fig:distn}
\end{figure}

In this study, we used a publicly available labelled DR fundus image dataset from the Asia Pacific Tele-Ophthalmology Society (APTOS)\footnote[1]{\href{https://asiateleophth.org/}{\texttt{https://asiateleophth.org/}}} to train, validate and test the model. The dataset has 3662 DR fundus images categorised into a five-class scale of increasing severity, namely normal, mild, moderate, severe and proliferative \cite{gulshan,aao}. Figure \ref{fig:distn} shows the distribution of classes in the dataset. 
The plot shows that the data is imbalanced as the normal class has the highest number of instances. We resolve the class imbalance problem by incorporating weights of the classes into the loss function. In this case, the majority class was given a small weight while the minority classes were given larger weights. Intuitively, during training the model places more emphasis on the minority classes, giving higher penalties to misclassifications made by those minority classes, and less emphasis on the majority class \cite{buda,justin}.

\begin{figure}[h]
	\centering
	\framebox{
	\begin{minipage}[t]{.2\textwidth}
		\centering
		\vspace{0pt}
		\includegraphics[scale=0.2]{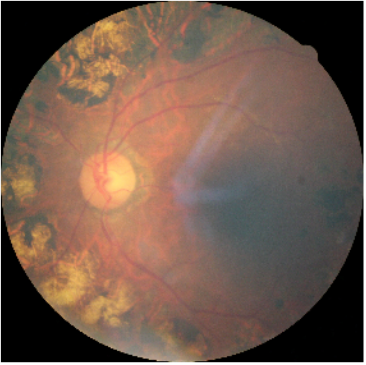}
		\label{fig: orig}
		\subcaption{\scriptsize Before CLAHE {  } }
	\end{minipage}
	\hspace{1ex}
	\begin{minipage}[t]{.2\textwidth}
		\centering
		\vspace{0pt}
		\includegraphics[width=2.5cm, height=2.7cm]{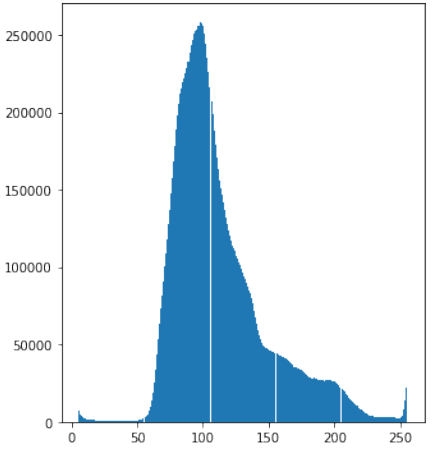}
		\vspace{-4pt}
		\label{fig: orig_distn}
		\subcaption{\centering \scriptsize Before CLAHE intensity distribution}
	\end{minipage}
	\hspace{7ex}
	\begin{minipage}[t]{.2\textwidth}
		\centering
		\vspace{0pt}
		\includegraphics[scale=0.2]{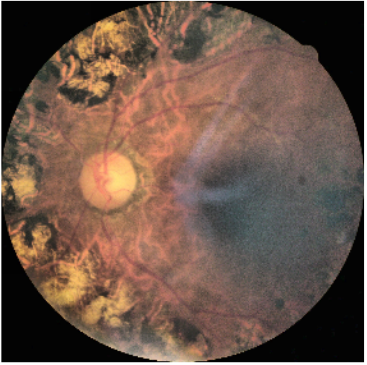}
		\label{fig: clahe}
		\subcaption{\scriptsize After CLAHE {   } }
	\end{minipage}
	\hspace{1ex}
	\begin{minipage}[t]{.2\textwidth}
		\centering
		\vspace{0pt}
		\includegraphics[width=2.5cm, height=2.7cm]{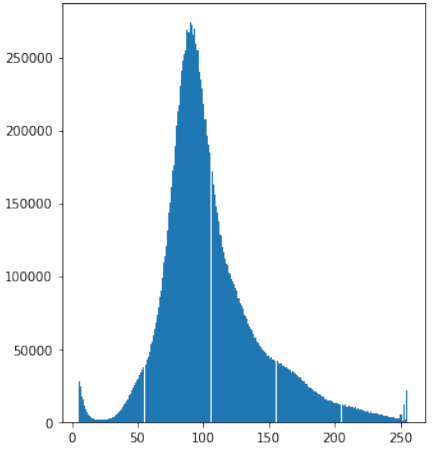}
		\label{fig: clahe_distn}
		\vspace{-4pt}
		\subcaption{\centering \scriptsize After CLAHE intensity distribution}
	\end{minipage}}
	\caption{Depicting a DR fundus image and its corresponding intensity distribution before and after applying CLAHE.}
	\label{fig: compare}
\end{figure}

For image pre-processing, we used both vertical and horizontal flips for training data augmentation as well as standard normalisation of the data. We also resized all of the images to either $224 \times 224$ or $299 \times 299$ depending on the pre-trained model we want to use. An important pre-processing technique used in this study is Contrast Limited Adaptive Histogram Equalization (CLAHE) \cite{zuiderveld} because DR fundus images often suffer from low contrast issues \cite{sahu}. Figure \ref{fig: compare} shows a comparison of the intensity distribution of a DR fundus image before and after applying CLAHE.
It can be seen that the DR fundus image before CLAHE has more Gaussian (white) noise \cite{sahu} making it potentially difficult to analyse. The Gaussian noise is reduced after adjusting the DR fundus image with CLAHE and its corresponding distribution is more normalised. This step is crucial because it can affect the sensitivity and specificity of the model \cite{sahu}.

The pre-processed images are passed to the model which in this case has a CNN backbone. We consider and compare four CNN backbones in this study, namely VGG16 \cite{simonyan}, ResNet50 \cite{resnet}, InceptionV3 \cite{szegedy} and InceptionResNetV2 \cite{inceptionv4}. The idea is to pass images through a pre-trained model and extract its feature maps at the last layer for classification. It should be noted that as images pass through a deep convolutional model, there exist trade-offs between losing spatial resolution and learning higher-level semantics, especially in the last convolutional layers. Neurons in those layers look for semantic class-specific information in the image, useful for discriminative purposes. Grad-CAM thus helps to interpret and explain individual components of a model \cite{zhou,selvaraju}. 

\begin{figure}[h]
	\centering
	\framebox{\includegraphics[width=\textwidth, height=4.6cm]{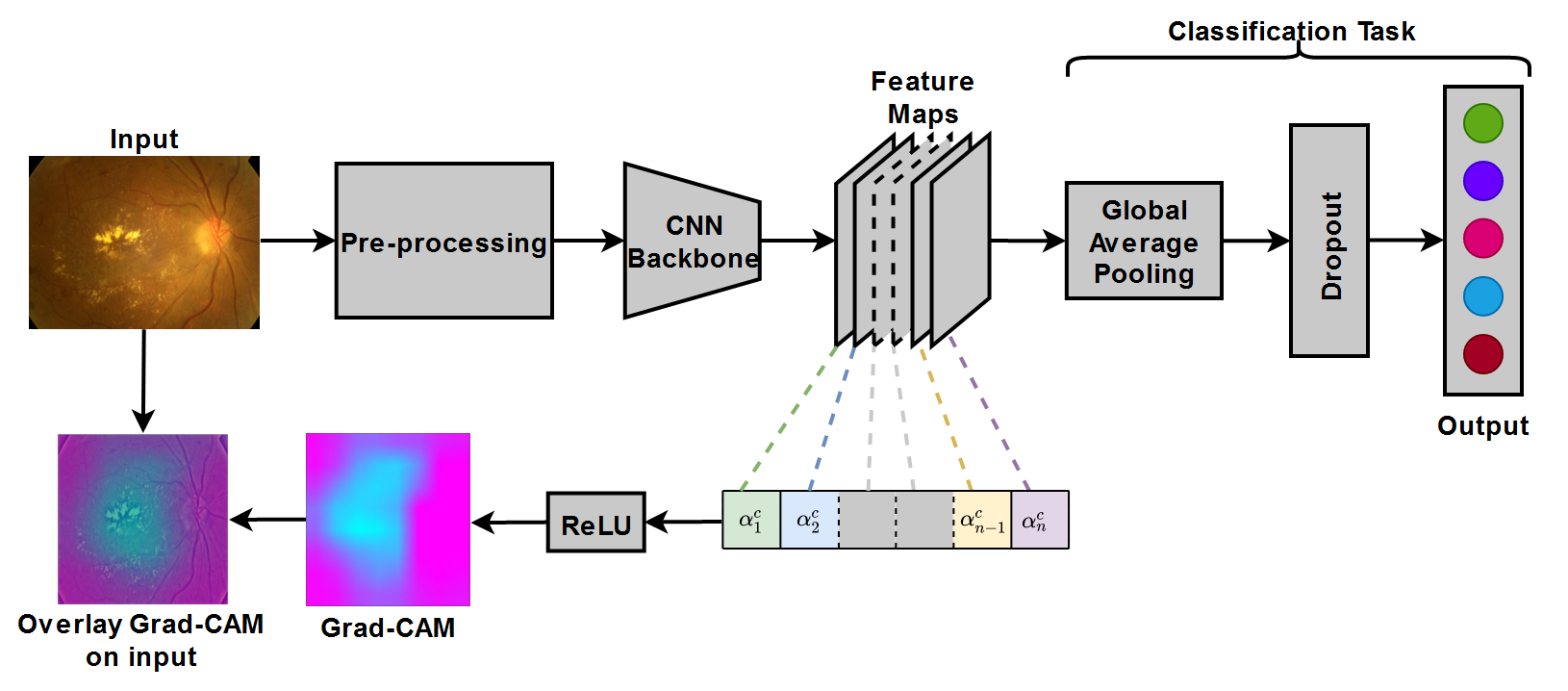}}
	\caption{Depicting the setup of generating a localisation map and classifying a DR fundus image.}
	\label{fig:workflow}
\end{figure}

Grad-CAM is a technique used to produce visual explanations from decisions made by a CNN. Specifically, it uses gradients of a class concept in the penultimate layer to produce a coarse localisation map which helps to identify discriminative regions in an image. For every class $c$, there are $k$ feature maps. To generate localisation maps, we first compute neuron importance weights $\alpha^c_k$ by global-average-pooling gradients of the target $y^c$ with respect to the feature maps $A^k$. It is important to note that $A^k$ is a spatial feature map hence it has width and height dimensions indexed by $i$ and $j$ \cite{selvaraju}. Neuron importance weight $\alpha^c_k$ is given by
\begin{equation}
\alpha^c_k = \overbrace{\frac{1}{Z} \sum_i \sum_j}^{\text{global average pooling}} \underbrace{\frac{\partial y^c}{\partial A^k_{ij}}}_{\text{gradients via backprop}}.
\end{equation}
Finally, we pass a linear combination of neuron importance weights $\alpha^c_k$ and feature maps $A^k$ through a ReLU function to generate the coarse localisation map. The generated map can then be overlaid on top of the input image to identify the region(s) of interest. Grad-CAM is given by
\begin{equation}
L^{c}_{\text{Grad-CAM}} = \text{ReLU}\underbrace{\bigg(\sum_k \alpha^c_k A^k\bigg)}_{\text{linear combination}}.
\end{equation}
Aside from using the feature maps to generate localisation maps, they are also used for discriminative purposes. In this study we replace the final layer with three successive layers, starting with a Global Average Pooling (GAP) layer, followed by a dropout (with 50\% probability) layer and finally a dense layer. In summary, we feed a CNN model with a preprocessed DR fundus image. We then generate localisation maps using feature maps extracted from the CNN model. In addition, the feature maps are used to classify the DR fundus images in an image-level manner. Figure \ref{fig:workflow} shows a diagramme of this approach. 
\FloatBarrier
\FloatBarrier
\section{Results}\label{s3}
In this section we evaluate the classification performance of the models and their ability to generate localization maps for DR fundus images. Models are trained using Keras with TensorFlow backend on an NVIDIA Tesla V100 GPU (available on the Centre for High Performance Computing \cite{chpc}) for 15 epochs. We use the sigmoid activation function for the last layer and cross entropy as the loss function for training. 

In this context, we use accuracy and Area Under Curve (AUC) as performance metrics over the test set. For every image we first predict a class and then generate the localisation map over the image. Table \ref{tab: performance} shows the performance of the classification task.
\begin{table}[h]
		\centering
		\caption{Performance Table for the Models}
		\label{tab: performance}
	\begin{tabular}{lcccccc}
		\hline
		\multicolumn{1}{c}{\multirow{2}{*}{Model}} & \multirow{2}{*}{Accuracy (\%)} & \multicolumn{5}{c}{AUC}\\ \cline{3-7}\multicolumn{1}{c}{}& & Normal & Mild & Moderate & Severe & Proliferative \\ \hline
		VGG16& 95.31&0.97&0.64&0.77&0.62&0.78\\ \hline
		ResNet50&94.56&0.97&0.72&0.82&0.71&0.75\\ \hline
		InceptionV3&96.07&0.97&0.67&0.84&0.62& 0.67\\ \hline
		InceptionResNetV2&94.39&0.96&0.77&0.81&0.68&0.69\\ \hline
	\end{tabular}
\end{table}
We observe that InceptionV3 performed better than the other models as it had the highest accuracy of 96.05\%. This is followed by VGG16, InceptionResNetV2 and ResNet50 in that order, however their performances are close to one other.

\begin{figure}[ht]
	\centering
	\framebox{
		\begin{minipage}[t]{.3\textwidth}
			\centering
			\includegraphics[scale=0.3]{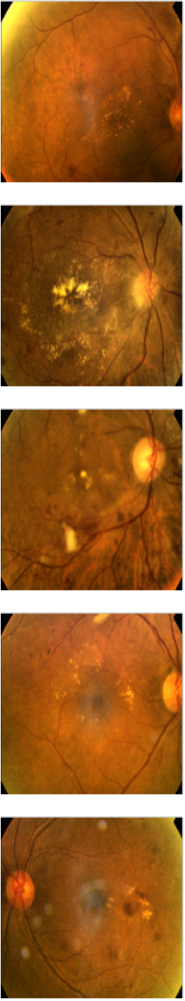}
			\subcaption{Input}
			\label{fig: input}
		\end{minipage}
		\begin{minipage}[t]{.3\textwidth}
			\centering
			\includegraphics[scale=0.3]{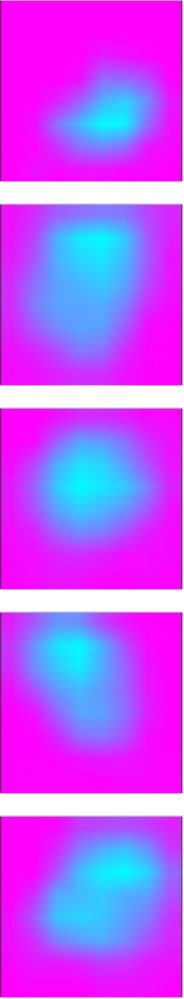}
			\subcaption{Grad-CAM}
			\label{fig: local}
		\end{minipage}
		\begin{minipage}[t]{.3\textwidth}
			\centering
			\includegraphics[scale=0.3]{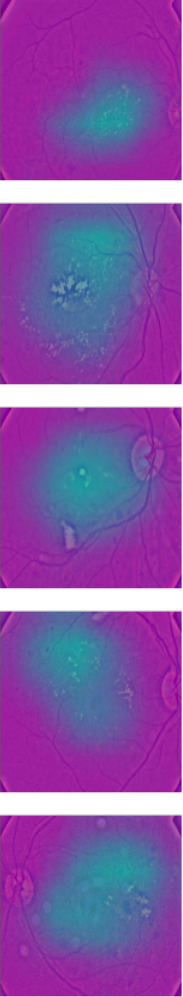}
			\subcaption{Overlaid Grad-CAM}
			\label{fig: over}
		\end{minipage}
	}
\caption{Generated Grad-CAM for higher-level semantics and overlaid Grad-CAM on randomly selected input images.}
\label{fig: final}
\end{figure}

Finally, we randomly select some input images (see Figure \ref{fig: input}) for demonstrative purposes. We pass the selected images through the InceptionV3 model (since it had the highest accuracy) to generate localisation maps and overlay the generated maps on the input images, as shown in Figure \ref{fig: final}. We observe that the model generates a blurry image in Figure \ref{fig: local}. The blurry effect can be attributed to the typical loss of spatial resolution as images pass through deep convolutional models. The blurry image in this case is the result of Grad-CAM. It may not be as useful until we map it on the input images. We see in Figure \ref{fig: over} that the Grad-CAM has highlighted regions of interest on the input images after mapping.
\FloatBarrier
\FloatBarrier
\section{Conclusions}\label{s4}
\vspace*{-0.65cm}
In this work we presented a technique which identifies regions of interest in DR fundus images and produces visual explanations for models. This could aid ophthalmologist in understanding the reasoning behind a model's discriminative process and speed up diagnosis. We observe high performance for classification as well as potentially useful localisation maps. We observed that the performance of different models are very close to each other. In future, we intend to use in-training attention mechanism such as stand-alone self-attention \cite{niki} to classify and generate localisation maps. This is because in-training attention mechanisms overcome the limitation of losing spatial resolution. 

\section*{Acknowledgement}
We would like to thank German Academic Exchange Service (DAAD) for kindly offering financial support for this research. Also, we thank the Centre for High Performance Computing (CHPC) \cite{chpc} for providing us with computing resource for this research. 

%
%

\end{document}